\documentclass[runningheads]{llncs}
\usepackage{graphicx}
\usepackage{comment}
\usepackage{amsmath,amssymb}
\usepackage{color}
\usepackage[pagebackref=true,breaklinks=true,colorlinks,bookmarks=false]{hyperref}
\usepackage[width=122mm,left=12mm,paperwidth=146mm,height=193mm,top=12mm,paperheight=217mm]{geometry}
\begin{document}
\makeatletter
\renewcommand{\paragraph}{%
  \@startsection{paragraph}{4}%
  {\z@}{1.0ex \@plus 0.5ex \@minus .5ex}{-1em}%
  {\normalfont\normalsize\bfseries}%
}
\makeatother
\pagestyle{headings}
\mainmatter
\title{Neural Object Learning for 6D Pose Estimation Using a Few Cluttered Images} 
\titlerunning{Neural Object Learning}

\author{Kiru Park \and
Timothy Patten \and
Markus Vincze}

\authorrunning{K. Park et al.}

\institute{Vision for Robotics Group \\
Automation and Control Institute, TU Wien, Vienna, Austria
\email{\{park,patten,vincze\}@acin.tuwien.ac.at}}

\maketitle

\begin{abstract}
Recent methods for 6D pose estimation of objects assume either textured 3D models or real images that cover the entire range of target poses. However, it is difficult to obtain textured 3D models and annotate the poses of objects in real scenarios. This paper proposes a method, Neural Object Learning (NOL), that creates synthetic images of objects in arbitrary poses by combining only a few observations from cluttered images. A novel refinement step is proposed to align inaccurate poses of objects in source images, which results in better quality images. Evaluations performed on two public datasets show that the rendered images created by NOL lead to state-of-the-art performance in comparison to methods that use 13 times the number of real images. Evaluations on our new dataset show multiple objects can be trained and recognized simultaneously using a sequence of a fixed scene. Our code\footnote{\url{https://github.com/kirumang/NOL}} and dataset\footnote{\url{https://www.acin.tuwien.ac.at/en/vision-for-robotics/software-tools/smot/}} are publicly available. The supplementary video summarizes the method and results (\url{https://youtu.be/fQJPS01cmac}). 
\end{abstract}

\section{Introduction} \label{introduction}

The pose of an object is important information as it enables augmentation reality applications by displaying contents in correct locations and robots to grasp and place an object precisely. Recently, learned features from color images using Convolutional Neural Networks (CNN) have increased performance of object recognition tasks~\cite{he2017mask,he2016deep_resnet} including pose estimation~\cite{kehl2017ssd,Li2019_CDPN,Park2019_Pix2Pose,rad2017bb8,Sundermeyer_2018_ECCV_implicit,Tekin_2018_CVPR,Zakharov2019_DPOD}. These methods have achieved the best performance on benchmarks for pose estimation using household objects~\cite{linemode_hinterstoisser2012,Kaskman_2019_ICCV_HB,xiang2017posecnn}. 

However, methods using CNNs require a large number of training images to cover potential view points of target objects in test environments. There have been two approaches to create training images for pose estimation methods: rendering synthetic images using textured 3D models~\cite{kehl2017ssd,Sundermeyer_2018_ECCV_implicit} or cropping real images and pasting them on random background images~\cite{cnn_pose:brachmann2016uncertainty_only_rgb,Park2019_Pix2Pose,rad2017bb8}. Recently, state-of-the-art performance has been accomplished by using both synthetic and real images~\cite{Li_2018_DeepIM,Li2019_CDPN,Peng_2019_CVPR_PVNet}. Unfortunately, both textured 3D models and large numbers of real images are difficult to obtain in the real world. Textured 3D models included in pose benchmarks are created with special scanning devices, such as the BigBIRD Object Scanning Rig~\cite{alli2015YCB} or a commercial 3D scanner~\cite{Kaskman_2019_ICCV_HB}. During this scanning operation, objects are usually placed alone with a simple background and consistent lighting condition. Precise camera poses are obtained using visible markers or multiple cameras with known extrinsic parameters. In the real environment, however, target objects are placed in a cluttered scene, are often occluded by other objects and the camera pose is imprecise. Furthermore, the manual annotation of 6D poses is difficult and time consuming because it requires the association between 3D points and 2D image pixels to be known. Thus, it is beneficial to minimize the number of images with pose annotations that are required to train 6D pose estimators. This motivates us to develop a new approach to create images of objects from arbitrary view points using a small number of cluttered images for the purpose of training object detectors and pose estimators.

\begin{figure}[t]
\centering
\includegraphics[width=\linewidth]{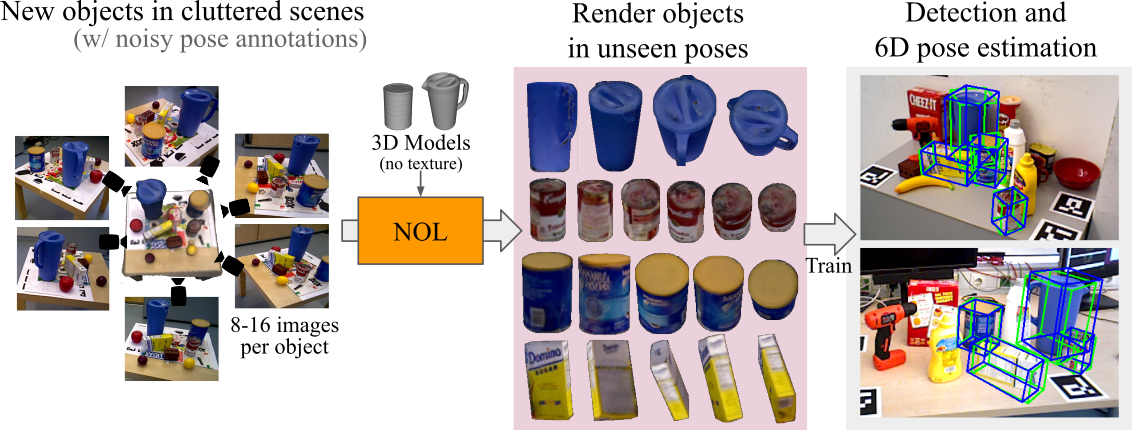}

\caption{The proposed method, NOL, uses a few cluttered scenes that consist of new target objects of which to render images in arbitrary poses. Rendered images are used to train pipelines for 6D pose estimation and 2D detection of objects}
\label{fig:coverimg}
\end{figure}

In this paper, we propose Neural Object Learning~(NOL), a method to synthesize images of an object in arbitrary poses using a few cluttered images with pose annotations and a non-textured 3D model of the object. For new objects, NOL requires 3D models and cluttered color images (less than 16 images in our evaluations) with pose annotations to map color information to vertices. To overcome pose annotation errors of source images, a novel refinement step is proposed to adjust poses of objects in the source images. Evaluation results show that images created by NOL are sufficient to train CNN-based pose estimation methods and achieves state-of-the-art performance. 

In summary, this paper provides the following contributions: \textbf{(1)}  Neural Object Learning that uses non-textured 3D models and a few cluttered images of objects to render synthetic images of objects in arbitrary poses without re-training the network. \textbf{(2)} A novel refinement step to adjust annotated poses of an object in source images to project features correctly in a desired pose without using depth images. \textbf{(3)} A new challenging dataset, Single sequence-Multi Objects Training (SMOT), that consists of two sequences for training and eleven sequences for evaluation, collected by a mobile robot, which represents a practical scenario of collecting training images of new objects in the real world. The dataset is available online. \textbf{(4)} Evaluation results that show images rendered by NOL, which uses 8 to 16 cluttered images per object, are sufficient to train 6D pose estimators with  state-of-the-art performance in comparison to methods that use textured 3D models and 13 times the number of real images. 

The remainder of the paper is organized as follows. Related work is briefly introduced in Sec~\ref{related_work}. The detailed description of NOL and the rendering process are described in Sec~\ref{method}. We report results of evaluations in Sec~\ref{evaluation} and analyze effects of each component in Sec~\ref{ablation}. Lastly, we conclude the paper in Sec~\ref{conclusion}.

\section{Related Work}\label{related_work}

In this section, we briefly review the types of training images used to train 6D pose estimation methods. The approaches that create a 3D model and map a texture of a novel object are discussed. The recent achievements of differentiable rendering pipelines are also introduced.

\subsection{Training Samples for 6D Pose Estimation}
    Previous work using CNNs requires a large number of training images of an object that covers a range of poses in test scenes sufficiently. Since it is difficult to annotate 6D poses of objects manually, synthetic training images are created using a textured 3D model of an object~\cite{kehl2017ssd}. However, it is difficult to obtain a 3D model with high quality texture from the real world without a special device, such as the BigBIRD Object Scanning Rig~\cite{alli2015YCB}, since precise pose information is required to correctly align texture images from different views. Using synthetic data introduces the domain gap between synthetic and real images, which should be specially treated with domain adaptation techniques~\cite{Zakharov2019_DPOD}. It is possible to use only approximately 200 real images and apply various augmentation methods to successfully train pose estimation pipelines~\cite{cnn_pose:brachmann2016uncertainty_only_rgb,Park2019_Pix2Pose,rad2017bb8}. However, the performance highly depends on the range of poses in the real training samples. As discussed in~\cite{Park2019_Pix2Pose}, the limited coverage of poses in training images causes inaccurate results for novel poses. To overcome this limitation, both real images and synthetic images are used for training~\cite{Li_2018_DeepIM,Li2019_CDPN,Peng_2019_CVPR_PVNet,Zakharov2019_DPOD}, which currently achieves state-of-the-art performance. The advantage of using both sources is that synthetic images supplement images for novel poses that are not observed in real images while real images regularize the network from over-fitting to synthetic images.

    However, both textured 3D models and more than 200 real images with pose annotations are difficult to obtain from the real world. Furthermore, textured 3D models in public datasets are captured separately from constrained environments~\cite{alli2015YCB,linemode_hinterstoisser2012,rgbddataset:tless,Kaskman_2019_ICCV_HB} such as single objects with a simple background and precise camera pose localization tools. However, this well-constrained setup is difficult to replicate in real scenarios, e.g., a target object on a table is often occluded by other objects, camera poses are noisy without manual adjustments, and lighting conditions are not consistent. Thus, it is challenging to derive training images of new objects from cluttered scenes. 

\subsection{3D Object Modeling and Multi-View Texturing}
    RGB-D images have been used to build 3D models by presenting an object in front of a fixed camera while rotating a turn table~\cite{v4r:rtmt_prankl2015rgb}, manipulating the object using a robot end-effector~\cite{Krainin_robot_hand_modeling_IJRR} or human hands~\cite{Wang2019_Modelling}. Alternatively, a mobile robot is used to actively move a camera to build a model of a fixed object in~\cite{thomas2017autonomous}. Even though these methods produce good 3D models in terms of geometry, textures are not optimized or even explicitly considered. Depth images are also required to align different views.
    
    On the other hand, it is possible to map multiple images from different views to 3D mesh models using camera pose information~\cite{fu2018_g2ltex,Waechter2014_MVS_texturing}. These approaches produce 3D models with high-quality textures since their optimization tries to assign continuous source images to neighboring pixels. However, these methods requires depth images for correcting pose errors, which causes misalignment of color values and disconnected boundaries when different source images are not correctly aligned. Image based rendering (IBR) has been used to complete a large scene by in-painting occluded area using multiple images from different view points~\cite{philip2018plane_IBR,shum2000review_ibr,thonat2016multi_IBR,whyte2009get_out}. These methods re-project source images to a target image using the relative poses of view points. Then, projected images from different views are integrated with a weighted summation or an optimization based on different objective functions. However, IBR methods are designed to complete large-scale scenes and suffer from noisy estimation of the camera and object poses, which causes blurry or misaligned images.

\subsection{Differentiable Rendering}
    The recent development of differentiable rendering pipelines enables the rendering process to be included during network training~\cite{Henderson2019_dirt,kato2018renderer,Justus_2019_deferred_rendering}. Therefore, the relationship between the 3D locations of each vertex and UV coordinates for textures are directly associated with pixel values of 2D rendered images, which have been used to create 3D meshes from a single 2D image~\cite{Henderson2019_dirt,kato2018renderer}. Furthermore, it is possible to render trainable features of projected vertices, which can be trained to minimize loss functions defined in a 2D image space.

The purpose of NOL is to generate synthetic images to train CNN-based pose estimators for new objects while minimizing the effort for obtaining training data from real applications. The knowledge of 3D representation and a few observations of objects are usually sufficient for humans to recognize objects in a new environment. Likewise, NOL composes appearances of objects in arbitrary poses using 3D models and a few cluttered and unconstrained images without using depth images, which is sufficient to train a pose estimator and achieve state-of-the-art performance.

\section{Neural Object Learning}\label{method}

    The objective of NOL is to create an image $X^{D}$ of an object in a desired pose $T^{D}$ using K source images, $\{I^1, I^k ... I^{K}\}$, with pose annotations, $\{T^1, T^k ... T^{K}\}$, and object masks that indicate whether each pixel belongs to the object or not in a source image, $\{M^1, M^k ... M^{K}\}$. We do not assume $T^k$ or $M^k$ to be accurate, which is common if the source images are collected without strong supervision such as marker-based localization or human annotation. 

\begin{figure}[t]
\centering
\includegraphics[width=\linewidth]{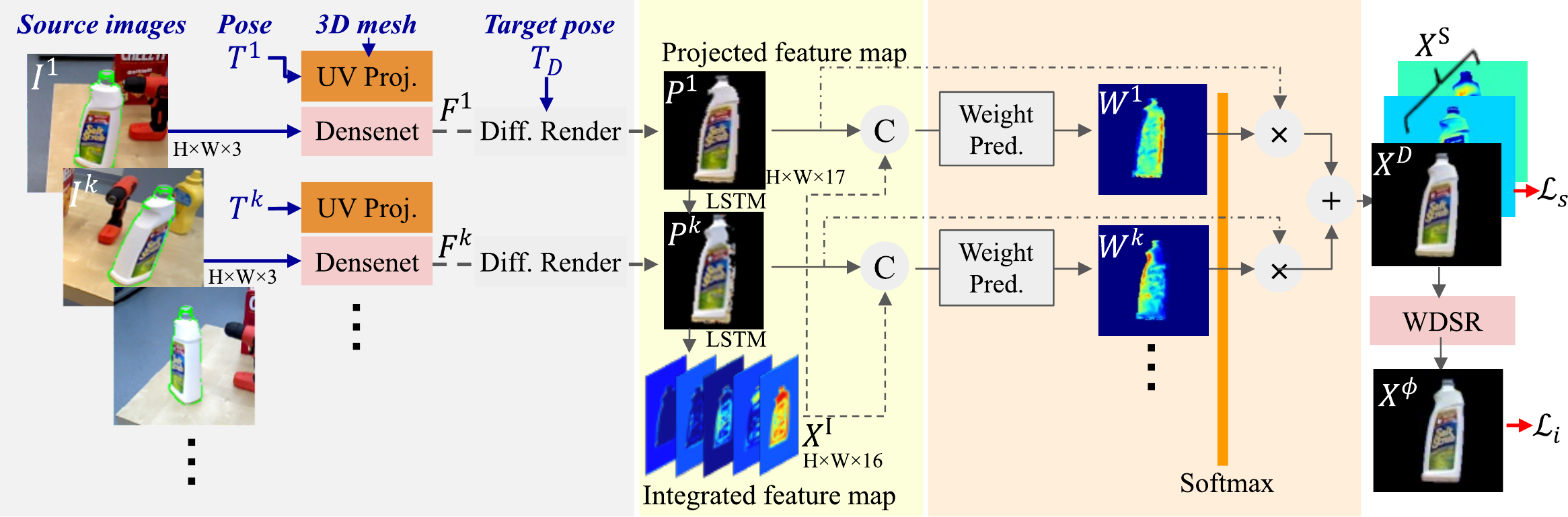}
\caption{An overview of the NOL architecture. $X^D$ is used as a rendered output while $X^\phi$ is used to compute the image loss for training}
\label{fig:arch1}
\end{figure}

\subsection{Network Architecture}
Fig.~\ref{fig:arch1} depicts an overview of the network architecture. Firstly, source images are encoded and projected to compute an integrated feature map in a target pose. Secondly, the weighted sum of projected feature maps are computed by predicting weight maps. A decoder block produces a decoded image that is used to compute the image loss. 

\paragraph{Integrated Feature Maps}  Each source image, $I^k \in H\times W \times 3$, is encoded with a backbone network, Densenet-121~\cite{Huang_2017_CVPR_dense_net}, to build feature pyramids using the outputs of the first four blocks. Each feature map from the pyramids is processed with a convolutional layer with 3$\times$3 kernels to reduce the number of an output channel of each block to 4, 3, 3, and 3. Each feature map is resized to the size of the original input using bi-linear interpolation. In addition to the feature map with 13 channels, original color images (3 channels) and face angles with respect to camera views (one channel) are concatenated. As a result, the encoded feature map $F^k$ of each input image $I^k$ has 17 channels. The UV coordinates of each vertex are computed using 2D projected locations of each visible vertex in an input pose $T^k$. These UV coordinates are then projected to the target pose $T^{D}$ using a differentiable renderer proposed in~\cite{Henderson2019_dirt}. Feature values of each pixel in a projected feature map $P^{k}$ are computed using bi-linear interpolation of surrounding feature values obtained from corresponding pixels from the encoded feature map $F^k$, which is similar to rendering an object with a separate texture image. The projected feature maps $P^{k \in K}$ are compiled by convolutional Long-Term and Short-Term memory (LSTM) layers to compute the integrated feature map $X^\textrm{I}$ with 16 output channels at the same resolution of the projected feature maps. This LSTM layer enables the network to learn how to extract valuable pixels from different source images while ignoring outlier pixels caused by pose errors. It is also possible to use a different number of source images without changing the network architecture. 
 
 \paragraph{Weight Prediction Block} The integrated feature map $X^\textrm{I}$ is concatenated with each projected feature map $P^{k}$ to compute a corresponding weight map $W^k$ in the weight prediction block, which implicitly encodes distances between $P^{k}$ and $X^\textrm{I}$ per pixel. The resulting weight maps $W^{k \in K}$ are normalized with the {\it Softmax} activation over K projected images. Therefore, the summation of the weighted maps over $W^{k \in K}$ is normalized for each pixel while keeping strong weights on pixels that have remarkably higher weights than others. The weighted sum of projected feature maps using the predicted weight maps produces the weighted feature map $X^\textrm{S}$. Since the first three channels of $P^k$ represent projected color values from source images, the first three channels of $X^\textrm{S}$ are a color image, which is referred to as the weighted rendering $X^{D}$ and the output image of NOL during inference. 
 
 \paragraph{Decoder Block} Since $X^{D}$ is obtained by the weighted summation of projected images, color values of pixels $X^{D}$ are limited to the color range of projected pixels. However, when training the network, the color levels of the source images can be biased by applying randomized color augmentations while maintaining the original colors for the target image. This causes the weight prediction block to be over-penalized even though color levels of $X^{D}$ are well balanced for given source images. This motivates us to add a module to compensate for these biased errors implicitly during training. An architecture used for the image super-resolution task, WDSR~\cite{yu2018wide_WDSR}, is employed as a decoder block to predict the decoded rendering $X^\phi$ in order to compute the losses during training. A detailed analysis regarding the role of the decoder is presented in Sec.~\ref{ablation}.

\subsection{Training}

The objective of training consists of two components. The first component, the {\it image loss}, renders a correct image $\mathcal{L}_\textrm{i}$ in a target pose. The second component, the {\it smooth loss}, minimizes the high frequency noise of the resulting images $\mathcal{L}_\textrm{s}$. 

\paragraph{Image Loss} The image loss $\mathcal{L}_\textrm{i}$ computes the difference between a target image $X^\textrm{GT}$ in a target pose and the decoded output $X^\phi$. In addition to the standard L1 distance of each color channel, the feature reconstruction loss~\cite{Justin_2016_perceptual} is applied to guide the predicted images to be perceptually similar to the target image as formulated by 
\begin{equation} \label{eq:image_loss}    
\mathcal{L}_\textrm{i} = \frac{1}{M^D} \sum_{p \in M^D}{ \lambda_i \lvert X^{\phi}_p- X^{GT}_p\lvert_1 + \lambda_f \lvert \psi(X^{\phi}_p)- \psi(X^{GT}_p)\lvert_1 },
\end{equation} 
where $M^D$ is a binary mask that indicates whether each pixel has at least a valid projected value from any input $I^{k \in K}$, and $\psi(\cdot)$ denotes outputs of a backbone network with respect to the image. The outputs of the first two blocks of DenseNet~\cite{Wang2019_DenseFusion} are used for the feature reconstruction loss. The parameters $\lambda_i$ and $\lambda_f$ are used to balance the losses. 

\paragraph{Smooth Loss} Even if the objective function in Eq.~\ref{eq:image_loss} guides the network to reconstruct the image accurately, the penalty is not strong when the computed image has high frequency noise. This is the reason why IBR and image in-painting methods~\cite{philip2018plane_IBR,thonat2016multi_IBR,whyte2009get_out} usually employ a smooth term. This minimizes the gradient changes of neighboring pixels even if pixel values are obtained from different source images. Similarly, we add a loss function to ensure smooth transitions for neighboring pixels in terms of color values as well as encoded feature values. This is formulated as
\begin{equation} \label{eq:smooth_loss}    
\mathcal{L}_\textrm{s} = \frac{\lambda_s}{M^D} \sum_{p \in M^D}{\nabla^2 X^{S}_p}.
\end{equation} 
The loss function creates a penalty when the gradients of color and feature values of each pixel are inconsistent with those of neighboring pixels. In contrast to the image loss, the weighted feature map $X^{S}$ is used directly instead of using the decoded output $X^{\phi}$. Thus, the weight prediction block is strongly penalized when producing high frequency changes in the predicted weight maps $W^{k\in K}$ and the weighted feature map $X^{S}$. 

\paragraph{Training using Synthetic Images with Pose Errors}
Synthetic images are created to train the NOL network. 3D models from the YCB-Video dataset~\cite{xiang2017posecnn} are used while ignoring original textures and instead applying randomly sampled images from MS-COCO~\cite{lin2014mscoco} as textures. After sampling a 3D model and a texture image, 10 images are rendered as a batch set in different poses with random background images. During training, one image from a batch set is chosen as a target image $X^{GT}$ and its pose is set to a desired pose $T^{D}$, and the other K images are assigned as input images $I^{k\in K}$. To simulate different lighting conditions, color augmentations are applied to the input images while no augmentation is applied to the target image. Pose errors are also simulated by applying random perturbations to the actual poses $T^{k\in K}$ of the input images during training. As a result of the perturbations, vertices are projected to wrong 2D locations, which produces wrong UV coordinates per vertex and outlier pixels in projected feature maps at a desired pose. This pose augmentation forces the network to be robust to pose errors while attempting to predict an accurate image in the target pose. A total of 1,000 training sets, consisting of 10 images per set, are rendered for training. The same weights are used to render objects in all evaluations in the paper after training for 35 epochs. Detailed parameters used for data augmentation are listed in the supplementary material.

\subsection{Gradient Based Pose Refinement and Rendering}
\begin{figure}[t]
\centering
\includegraphics[width=\linewidth]{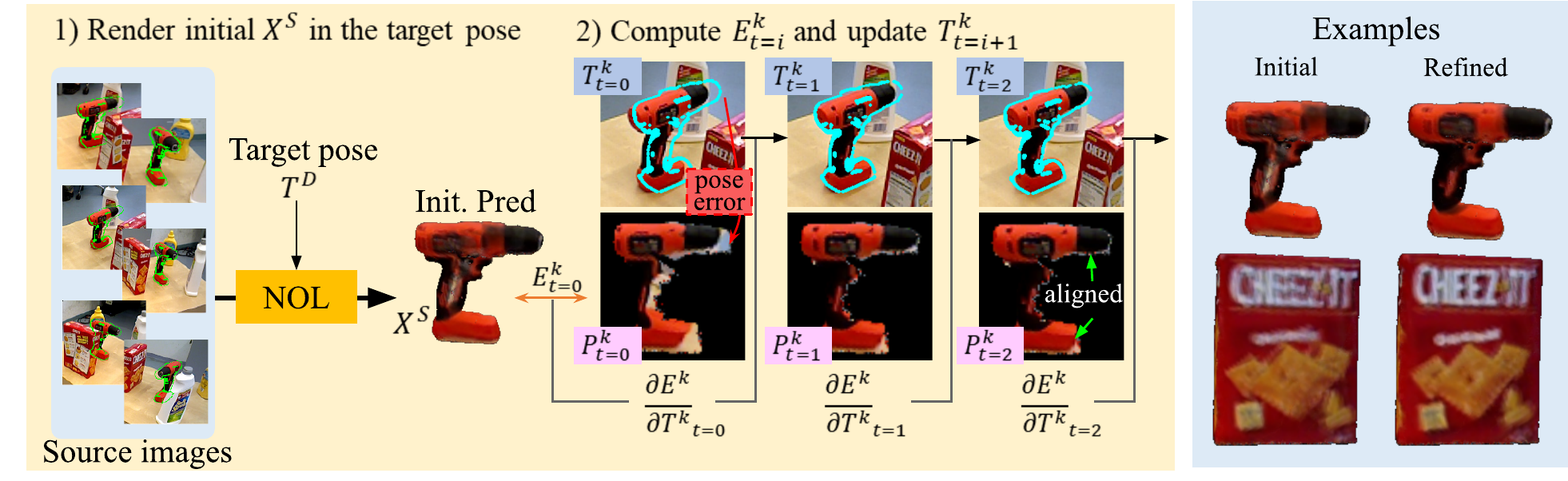}

\caption{An overview of the proposed pose refinement process and example results. The partial derivative of the projection error $E^k$ is used to update each input pose $T^k$}
\label{fig:refinement}
\end{figure}

The error in an input pose $T^k$ causes crucial outlier pixels in the projected feature map $P^{k}$ at the desired pose. Fig.~\ref{fig:refinement} shows an example of wrong pixels in the projected feature map obtained from the ground plane (blue) in the source image due to the error of the initial pose $T^k_{t=0}$. As discussed in Sec.~\ref{related_work}, the differentiable renderer enables derivatives of 3D vertices of a 3D model to be computed with respect to the error defined in 2D space. Since 3D locations of vertices and UV coordinates in the desired pose are derived by matrix multiplications, which is differentiable, the gradient of each input pose $T^k$ can be derived to specify a direction that decreases the difference between a desired feature map and each projected feature map $P^{k}$. The first prediction of NOL, $X^{S}_{t_0}$, without refinement is used as an initial desired target. The goal of the refinement step is to minimize the error, ${E^k}$, between the initial target and each projected feature map $P^{k}$. In every iteration, the partial derivative of the projection error ${E^k}$ with respect to each input pose $T^k_{t=t_i}$ is computed by
\begin{equation} \label{eq:refinement}    
\Delta T^k_{t_i} = \frac{\partial E^k}{\partial T^k_{t_i}}  = \frac{\partial \lvert X^{S}_{t_0}-P_{t_i}^{k}\lvert}{\partial T^k_{t_i}},
\end{equation}
and the input pose at the next iteration $T^k_{t_i+1}$ is updated with a learning step~$\delta$, i.e. $T^k_{t_i+1}  = T^k_{t_i} - \delta \Delta T^k_{t_i}$. In our implementation, translation components in $T^k$ are directly updated using $\Delta T^k_{t_i}$. On the other hand, updated values for rotation components, $\mathbb{R}^{3 \times 3}$, do not satisfy constraints for the special orthogonal group, $SO(3)$. Thus, the rotation component of $\Delta T^k_{t_i}$ is updated in the Euler representation and converted back to the rotation matrix. As depicted in Fig.~\ref{fig:refinement}, the iterations of the refinement step correctly remove the pose error so that the projected image no longer contains pixels from the background, which decreases blur and mismatched boundaries in the final renderings. After refining every input pose $T^k_{t=t_0}$ until the error does not decrease or the number of iterations exceeds 50, the final output $X^{D}$ is predicted using the refined poses $T^k_{t=t_f}$. 

\section{Evaluation}\label{evaluation}
This section presents the evaluation of the proposed NOL approach in relation to the task of 6D object pose estimation. We introduce datasets used in the evaluation and provide implementation details of NOL. The evaluation results show that the quality of NOL images are sufficient for pose estimation and leads to outperforming other methods trained with synthetic images using textured 3D models or real images.

\subsection{Datasets}
Three datasets are used for evaluation: LineMOD~\cite{linemode_hinterstoisser2012}, LineMOD-Occlusion (LineMOD-Occ)~\cite{brachmann2014learning_occlusion}, and a novel dataset created to reflect challenges of real environments. LineMOD and LineMOD-Occ have been used as standard benchmarks for 6D pose estimation of objects. LineMOD provides textured 3D models and 13 test sequences that have an annotated object per image. The 3D models are created by placing each object alone on a plane and performing a voxel-based 3D reconstruction~\cite{linemode_hinterstoisser2012}. LineMOD-Occ is created by additionally annotating eight objects presented in a test sequence in LineMOD. Previous works reporting results on these dataset have used either synthetic images using given 3D models~\cite{kehl2017ssd,Sundermeyer_2018_ECCV_implicit} or 15\% real images obtained from test sequences (183 images per object)~\cite{cnn_pose:brachmann2016uncertainty_only_rgb,Park2019_Pix2Pose,rad2017bb8,Tekin_2018_CVPR} for training. In contrast to previous work, images created by NOL are used to train both a pose estimator and a 2D detector.

\paragraph{SMOT}
A new dataset, Single sequence-Multi Objects Training, is created to reflect real noise when training images are collected from a real scenario, i.e. a mobile robot with a RGB-D camera collects a sequence of frames while driving around a table to learning multiple objects and tries to recognize objects in different locations. The dataset consists of two training sequences and eleven test sequences using eight target objects sampled from the YCB-Video~\cite{xiang2017posecnn} dataset. Two training sequences, that include four target objects per sequence, are collected by following trajectories around a small table. Camera poses of frames are self-annotated by a 3D reconstruction method~\cite{Zhou2018_open3d} while building a 3D mesh of the scene. 3D models provided in YCB-Video are aligned to the reconstructed mesh and corresponding object poses are computed using camera poses. No manual adjustment is performed to preserve errors of self-supervised annotations. On the other hand, test images are collected with visible markers to compute more accurate camera poses while moving the robot manually in front of different types of tables and a bookshelf. As a result, each object has approximately 2,100 test images. The supplementary material includes more details of the dataset.

\subsection{Implementation Details}
For the NOL network, the resolution of input and target images are set to 256$\times$256. A number of source images, $K$, is set to 8 for training and 6 for inference. The loss weights are set to, $\lambda_i$=5, $\lambda_f$=10, and $\lambda_s$=1. Detail parameters are reported in the supplementary material.

\paragraph{Sampling of Source Images} \label{sec:sampling}

To render NOL images for training a pose estimator, source images are sampled from the training sequences of a dataset. For LineMOD and LineMOD-Occ, a maximum of 16 images per object are sampled from the same training splits of real images used in previous work~\cite{cnn_pose:brachmann2016uncertainty_only_rgb,Park2019_Pix2Pose,rad2017bb8,Tekin_2018_CVPR}. Since objects are fully visible in the training set, images are simply sampled using pose annotations. In each sampling iteration, an image is randomly sampled and images that have similar poses (less than 300mm translation and 45$^\circ$ rotation), are removed. The sampling is terminated when no more images remain. In contrast to LineMOD and LineMOD-Occ, the visibility of each object varies in the training set of SMOT. In order to minimize the number of source images, a frame with the highest value is selected at each sampling iteration by counting the number of visible vertices that have not been observed in the previously sampled frames. The sampling iteration is terminated when no frame adds additional observed vertices.

\paragraph{Rendering NOL Images}
    Each target object is rendered using NOL in uniformly sampled poses defined over an upper-hemisphere for every 5$^\circ$ for both azimuth and elevation. For each target pose, 6 images are chosen from sampled images using the same image sampling procedure while limiting the target vertices to visible vertices in the pose. As a result, 1296 images are rendered. The in-plane rotation is applied to rendered images by rotating them from -45$^\circ$ to 45$^\circ$ for every 15$^\circ$. For LineMOD and LineMOD-Occ, synthetic images are also rendered in the same sampled poses using given 3D models to train a pose estimator for comparison. Rendering takes approximately 3.6 seconds per image, which consists of 19ms for the initial prediction, 3597ms for the pose refinement with maximum 25 iterations, and 19ms for the final prediction.
    
\paragraph{Training Recognizers}
   To show whether NOL images are sufficient to estimate poses of objects in arbitrary poses using a recent RGB-based pose estimation method,
   an official implementation of state of the art for 6D pose estimation using color images, Pix2Pose~\cite{Park2019_Pix2Pose}, is used. This method is one of the most recent methods~\cite{Li2019_CDPN,Park2019_Pix2Pose,Zakharov2019_DPOD} that predict objects' coordinate values per pixel. To increase the training speed and decrease the number of training parameters, the discriminator and the GAN loss are removed. All other aspects are kept the same except for the number of training iterations, which is set to approximately 14K because of the decreased number of trainable parameters. Resnet-50~\cite{he2016deep_resnet} is used as a backbone for the encoder and weights are initialized with pre-trained weights on ImageNet~\cite{cnn:imagenet}. Since the online augmentation of Pix2Pose uses cropped patches of objects on a black background, no further image is created for training. On the other hand, images that contain multiple objects are created to train a 2D detector by pasting masked objects onto random background images. A total of 150,000 training images are created and used to train Retinanet~\cite{Lin_2017_ICCV_retinanet} with Resnet-50 backbone.

\subsection{Metrics}

 The AD\{D$\vert$I\} score~\cite{linemode_hinterstoisser2012} is a common metric used to evaluate pose estimation. This metric computes the average distance of vertices between a ground truth pose and a predicted pose (ADD). For symmetric objects, distances to the nearest vertices are used (ADI). This metric has been widely used to compare the pose estimation performance of different methods. The predicted pose is regarded as correct if the average distance is less than 10\% of the diameter of each object.
 
 The recent challenge for pose benchmark~\cite{BOP2019} proposes a new metric that consists of three different pose errors (VSD, MSSD, and MSPD)~\cite{hodavn2016evaluation} and their average recall values over different thresholds. The metric is robust to pose ambiguities caused by occlusion. The metric is used to evaluate the performance on LineMOD-Occ because it is more suitable and also allows comparison with the results of the recent benchmark. 

\begin{figure}[t]
\centering
\includegraphics[width=0.95\linewidth]{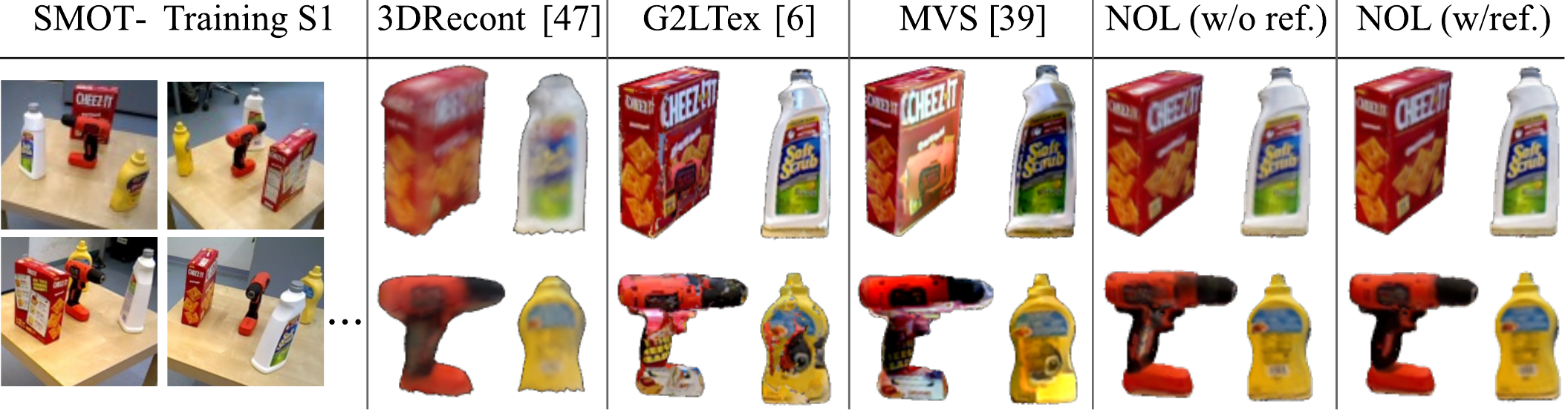}
\caption{Rendered results of SMOT objects using a training sequence. NOL successfully removes pixels from the background and other objects due to pose refinement}
\label{fig:render_ex}
\end{figure}

\subsection{Quality of Rendered Images}
Fig.~\ref{fig:render_ex} shows the rendered images using a training sequence of SMOT. Renderings in {\it 3DRecont} show object models extracted directly from a reconstructed 3D mesh of the training scene~\cite{Zhou2018_open3d}. Both MVS~\cite{Waechter2014_MVS_texturing} and G2LTex~\cite{fu2018_g2ltex} use the same images sampled for rendering NOL images with the same pose annotations. Multi-view texturing methods create less blurry textures than NOL for planar surfaces since they try to map an image to a large area without combining pixels from other images. However, this induces misaligned results when the input poses are inaccurate even if depth images are used to optimize poses as in~\cite{fu2018_g2ltex}, e.g. doubled letters on the cheeze-it box. On the other hand, results of NOL after pose refinement (last column) removes these doubled textures by correcting pose errors using color images only, which is robust to depth registration errors. Furthermore, NOL successfully rejects outlier pixels from other objects and the background.

\subsection{Pose Estimation: LineMOD}

\begin{figure}[t]
\centering
\includegraphics[width=\linewidth]{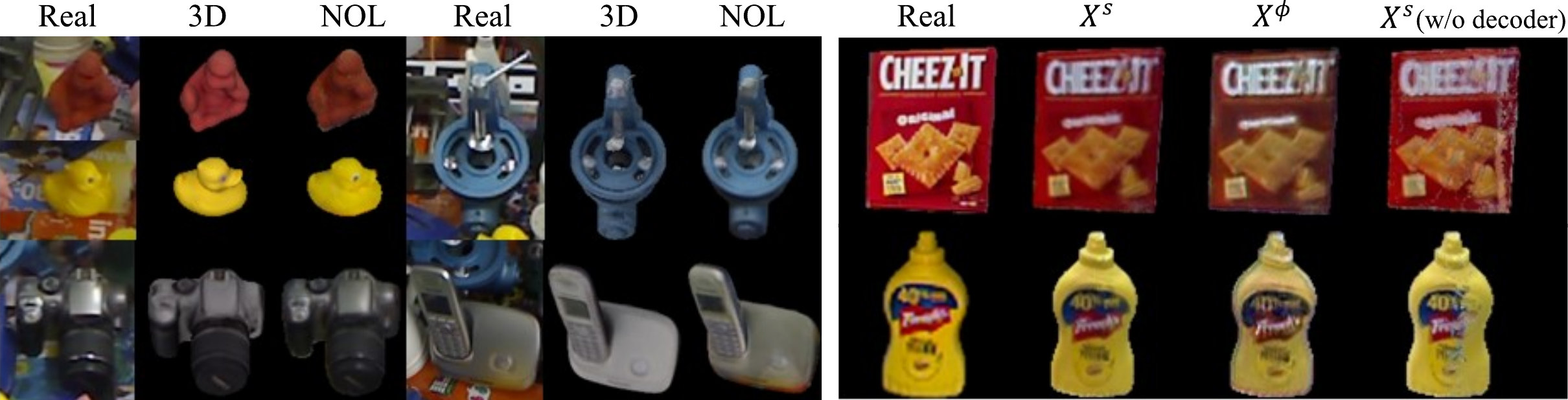}

\caption{Left: examples of rendered images of objects in LineMOD using NOL. Right: outputs of weighted renderings, decoded renderings, and weighted renderings after training without the decoder block}
\label{fig:linemod_rendering}
\end{figure}

\setlength{\tabcolsep}{3.5pt}
\begin{table}[t]
\begin{center}
\caption{Evaluation results on LineMOD. The ADD score is used except for {\it Eggbox} and {\it Glue} that use the ADI score}
\label{table:linemod}
\scriptsize
\begin{tabular}{c|| c c c c c c || c c c  |c c}

    Types & \multicolumn{11}{c}{Required data for training}\\
\hline
 Texture & -  & \checkmark & \checkmark & \checkmark & \checkmark &\checkmark   & -  & \checkmark & \checkmark & -&\checkmark\\
 
No. real images & \texttt{14}  & - & - & - & -&\texttt{183}   &\texttt{14}  & - & - & \texttt{183}&\texttt{183}\\
 
 GT 6D pose & \checkmark  & - & - & - & -&-    & \checkmark  & - & - & \checkmark&\checkmark\\

  Depth image & -  & - & - & - & -&\checkmark  & -  & - & - & \checkmark&\checkmark\\
\hline

 Test type  &  \multicolumn{6}{c||}{RGB w/o refinement} & \multicolumn{3}{c|}{RGB + ICP}& \multicolumn{2}{c}{RGB-D} \\
\hline
Training on & \multicolumn{1}{c|}{\textbf{NOL}} & \multicolumn{4}{c|}{synthetic images}&{real+syn} &\multicolumn{1}{c|}{\textbf{NOL}} &\multicolumn{2}{c|}{syn}&\multicolumn{1}{c|}{real}&\multicolumn{1}{c}{r+s} \\
\hline
Method & \cite{Park2019_Pix2Pose} & \cite{Park2019_Pix2Pose}  & \cite{kehl2017ssd} &  \cite{Sundermeyer_2018_ECCV_implicit} &\cite{Zakharov2019_DPOD}& \cite{rad2018domain}&\cite{Park2019_Pix2Pose}&\cite{Park2019_Pix2Pose}&\cite{Sundermeyer_2018_ECCV_implicit}&\cite{Wang2019_DenseFusion}&\cite{he2019pvn3d}\\
\hline
Ape  & 35.4 & 10.0 &2.6 & 4.0 & \textbf{37.2} &  19.8 &\textbf{95.2} &92.7 &20.6 & 92.3& \textbf{97.3} \\
Benchvise  & 55.6 & 13.4 &15.1 & 20.9& 66.8& \textbf{69.0} &\textbf{99.0}&90.4 & 64.3 & 93.2& \textbf{99.7}\\
Camera  & 37.5 & 4.4 &6.1 & 30.5 & 24.2 &  \textbf{37.6}    &\textbf{96.6}&77.9 & 63.2 & 94.4& \textbf{99.6}\\
Can  & \textbf{65.5} & 26.4 &27.3 & 35.9 & 52.6 &  42.3     &\textbf{97.6}&85.8 & 76.1 & 93.1& \textbf{99.5}\\
Cat  & \textbf{38.1} & 24.8 &9.3 & 17.9 & 32.4 &  35.4      &\textbf{98.6}&90.1 & 72.0 & 96.5& \textbf{99.8}\\
Driller  & 52.2 &9.1 &12.0 & 24.0 & \textbf{66.6} &  54.7 &\textbf{98.0}&66.1 & 41.6 & 87.0& \textbf{99.3}\\
Duck  & 14.7 & 3.7&1.3& 4.9 & 26.1  &  \textbf{29.4}        & \textbf{89.1} &82.3 & 32.4 & 92.3& \textbf{98.2}\\
Eggbox  & \textbf{93.7} &34.6 &2.8 & 81.0 & 73.4 &  85.2    &\textbf{99.2}&88.7& 98.6 & 99.8& \textbf{99.8}\\
Glue  & 63.1 & 35.1 &3.4 & 45.5 & 75.0  & \textbf{77.8}     &\textbf{96.5}&92.2& 96.4 & \textbf{100}& \textbf{100}\\
H.puncher  & 34.4 & 3.7&3.1 & 17.6 &24.5 &  \textbf{36.0}   &\textbf{93.2}&46.3 & 49.9 & 92.1& \textbf{99.9}\\
Iron  & 57.9 & 30.4 &14.6 & 32.0 & \textbf{85.0}&  63.1     &\textbf{99.3}&93.5& 63.1 & 97.0& \textbf{99.7}\\
Lamp  & 54.2 &6.7 &11.4 & \textbf{60.5} & 57.3&  75.1      &\textbf{96.6}&39.3 & 91.7 & 95.3& \textbf{99.8}\\
Phone  & 41.8 & 13.8 &9.7 & 33.8 & 29.1 &  \textbf{44.8}    &\textbf{92.8}&79.1 & 71.0 & 92.8& \textbf{99.5}\\
\hline
Average  & 49.5 & 16.6 &9.1 & 28.7 & 50.0& \textbf{ 51.6}& \textbf{96.3}&78.8& 64.7 & 94.3& \textbf{99.4}\\

\end{tabular}
\end{center}
\end{table}
The left side of Table~\ref{table:linemod} shows the results when RGB images are used for pose estimation. Since no real image is directly cropped and used to train the pose estimator, the results are mainly compared against methods that use synthetic images only for training. The method trained with NOL images outperforms the same method trained with synthetic images using the given 3D models. This verifies that the quality of NOL images are more similar to appearances of real objects. The results of objects with metallic or shiny surfaces, e.g., {\it Camera}, {\it Phone} ,and {\it Can}, show significant improvements against other results obtained with synthetic training images without any real observation. As depicted in Fig.~\ref{fig:linemod_rendering}, NOL realizes the details of shiny and metallic materials by optimizing colors of each view separately. The performance is competitive to the best method that uses real color and depth images of objects for domain adaptation. 

 NOL images tend to contain noisy boundaries especially around the lower parts of objects where NOL mistakenly extracts pixels from the background table (see the bottom of {\it Phone} in Fig.~\ref{fig:linemod_rendering}). This limits the translation precision of predictions along the principle camera axis ($z$-axis). To decrease the translation errors, ICP refinement is applied to refine poses using depth images as reported on the right side of Table~\ref{table:linemod}. The method trained with NOL images outperforms state-of-the-art trained with synthetic images. The result is competitive to state-of-the-art results in the last two columns even though the methods~\cite{he2019pvn3d,Wang2019_DenseFusion} use more than 13 times the number of real images for training.
 
\subsection{Pose Estimation: LineMOD-Occ} 
  The same models used in the LineMOD evaluation are used to test on LineMOD-Occ as reported in Table~\ref{table:linemod-occ}. Similar to the LineMOD evaluation, methods trained by synthetic images are mainly compared. The evaluation protocol used in the recent pose challenge~\cite{BOP2019} is applied with the same test target images. The result of~\cite{Park2019_Pix2Pose} using synthetic images is obtained by re-training the network with Resnet-50 backbone, which performs better than the official result in the challenge~\cite{BOP2019}.
 
 The performance of this method is significantly improved by using images created by NOL for training with RGB inputs and with the inclusion of ICP refinement using depth images. Furthermore, using NOL images leads to the method outperforming state of the art using color images~\cite{Li2019_CDPN} and the best performing method on this dataset~\cite{vidal20186d}.

\setlength{\tabcolsep}{3pt}
\begin{table}[t]
\begin{center}
\caption{Evaluation results on LineMOD-Occ. The results of other methods are cited from the last 6D pose challenge~\cite{BOP2019}}
\label{table:linemod-occ}
\begin{tabular}{c|| c | c c c c || c| c| c|c}
 Type  &  \multicolumn{5}{c||}{RGB w/o refinement} & \multicolumn{3}{c|}{RGB+ICP3D}&Depth \\
\hline
 Train source &  \multicolumn{1}{c|}{\textbf{NOL}} & \multicolumn{4}{c||}{Syn. using 3D models} &  \multicolumn{1}{c|}{\textbf{NOL}} & \multicolumn{2}{c|}{\textrm{Syn}}&\multicolumn{1}{c}{3D model}\\
\hline
Method & \cite{Park2019_Pix2Pose} &\cite{Park2019_Pix2Pose} &\cite{Sundermeyer_2018_ECCV_implicit} & \cite{Li2019_CDPN} & \cite{Zakharov2019_DPOD} & \cite{Park2019_Pix2Pose}& \cite{Park2019_Pix2Pose} & \cite{Sundermeyer_2018_ECCV_implicit} & \cite{vidal20186d}\\

\hline
BOP Score  & \textbf{37.7} & 20.0 &14.6 & 37.4& 16.9 & \textbf{61.3} &45.3&23.7 & 58.2\\

\end{tabular}
\end{center}
\end{table}

\setlength{\tabcolsep}{3.0pt}
\begin{table}[t]
\begin{center}
\caption{Evaluation results on the SMOT dataset}
\label{table:smot}
\begin{tabular}{r|| c| c |c |c || c| c|c| c}

 Type  &  \multicolumn{4}{c||}{RGB} & \multicolumn{4}{c}{RGB-D (ICP3D)} \\
\hline
 3D Model  &  \multicolumn{3}{c|}{Precise} & \multicolumn{1}{c||}{Recont} &  \multicolumn{3}{c|}{Precise} & \multicolumn{1}{c}{Recont}  \\
\hline
Train source  & Real & G2Ltex & \textbf{NOL} & \textbf{NOL} & Real & G2Ltex & \textbf{NOL} & \textbf{NOL}\\
\hline
AD\{D$\vert$I\} score & 25.0 & 25.7 & \textbf{35.5} &  22.5 &  86.5 &  82.0 & \textbf{90.0} & 80.6  \\ 
\hline
mAP$_{IoU=50}$ & 88.3 & 90.2 & \textbf{90.7}&  73.9 &  - &  - & - & - \\ 
\end{tabular}
\end{center}
\end{table}

\subsection{Pose Estimation: SMOT} \label{smot_eval}
The pose estimator~\cite{Park2019_Pix2Pose} and the 2D detection method~\cite{Lin_2017_ICCV_retinanet} are trained using crops of entire real images where each object is visible more than 50\%. This is an average of 364 images per objects. For G2Ltex and NOL, up to 16 images per object are sampled as explained in Sec.~\ref{sec:sampling} to render training images.
 
 Table~\ref{table:smot} shows pose estimation and 2D detection results in terms of the AD\{D$\vert$I\} score and the mean Average Precision (mAP)~\cite{cnn:pascalvoc}. The results using NOL images outperforms other methods using real images and models textured by G2Ltex for both RGB and RGB-D inputs. This is because real images do not fully cover target poses and objects are often occluded by other objects in training images. The comparison with G2LTex provides a quantitative verification regarding the better quality of renderings created by NOL using the same source images. The results denoted with {\it Recont} are obtained using reconstructed 3D models instead of precise 3D models for rendering. The performance drops significantly since the NOL images are noisier and blurrier due to geometrical errors of models. This indicates that precise 3D models are important for NOL to generate high-quality images.

\section{Ablation Study} \label{ablation}
This section analyzes factors that influence the quality of NOL images. The perceptual similarity~\cite{Zhang_2018_CVPR} is used to measure quality of generated images in comparison to the real images. We sample 10 test images per object in SMOT (80 images), render the objects at GT poses, and compare them with real images. 

\begin{table}[t] 
\begin{center}
\caption{Perceptual similarity (smaller is better) of rendered images with different configurations}
\label{table:ablation_main}
\begin{tabular}{c||ccc|cccc}
 & \multicolumn{3}{c|}{Components} & \multicolumn{4}{c}{Loss functions} \\
 \hline
Setup & All & w/o Decoder & w/o LSTM & $\mathcal{L}_\textrm{1}$ (RGB) & $\mathcal{L}_\textrm{i}$ & $\mathcal{L}_\textrm{i}+\mathcal{L}_\textrm{s}$ & $ +\mathcal{L}_\textrm{GAN}$ \\
\hline
w/o Ref & \textbf{0.181} & 0.289 & 0.247  & 0.194 & 0.184 & \textbf{0.181} & 0.188\\
w/ Ref &\textbf{0.173} & 0.279 & 0.241 & 0.184 & 0.177 & \textbf{0.173} & 0.183

\end{tabular}
\end{center}
\end{table}

\paragraph{Components} Table~\ref{table:ablation_main} shows the most significant improvement comes from the decoder. The right side of Fig.~\ref{fig:linemod_rendering} shows qualitative results of weighted renderings $X^{D}$, decoded renderings $X^{\phi}$, and results after training the network without the decoder. As discussed in Sec.~\ref{method}, the network trained with the decoder converges to produce $X^{D}$ in a neutral color level as a reference image while the decoder absorbs over-penalized errors caused by randomly biased colors. The results denoted as \textit{w/o LSTM} are derived by replacing the LSTM module with an simple average over projected features $P^{k}$. In this case, the results drops significantly since the LSTM module highlights valuable pixels among projected pixels. The refinement step consistently improves the image quality for all configurations.

\paragraph{Loss Function} The best results are made with all proposed losses $\mathcal{L}_\textrm{i}$+$\mathcal{L}_\textrm{s}$. The perceptual loss in addition to the standard L1 loss significantly improves the performance by guiding the network to preserve perceptual details, like edges, with less blurry images while the smooth loss $\mathcal{L}_\textrm{s}$ additionally reduces the high-frequency noise. As the adversarial loss~\cite{goodfellow2014generative} provides better performance for image reconstruction tasks, the adversarial loss $\mathcal{L}_\textrm{GAN}$ is added to our loss function, which does not improve the result in our implementation. 

\begin{figure}[hbt]
\centering
\includegraphics[width=\linewidth]{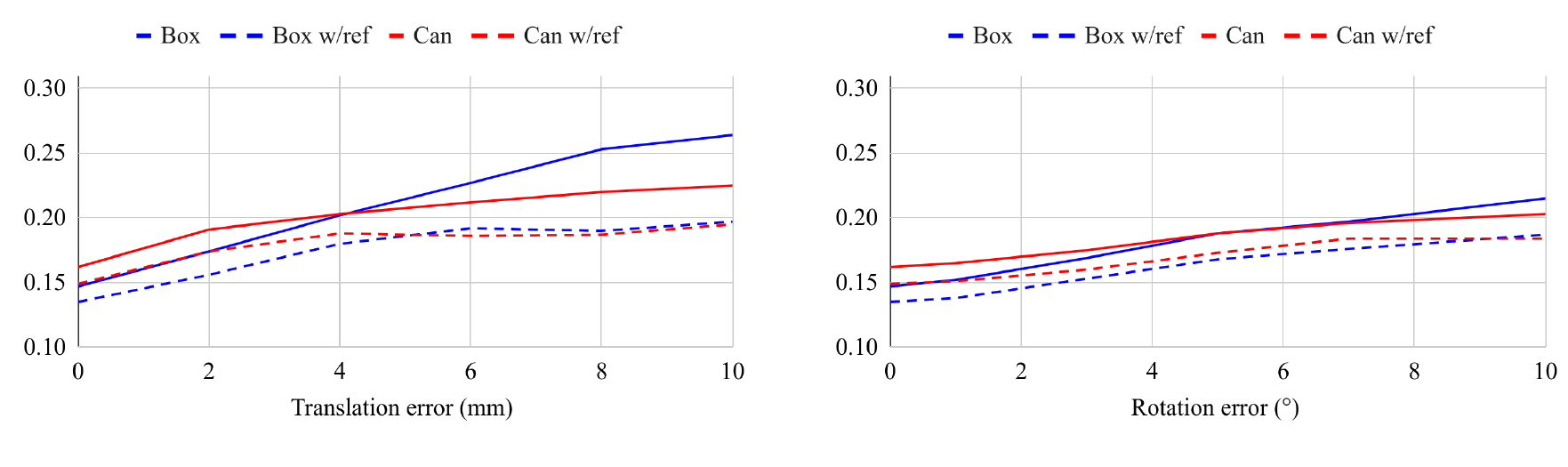}
\caption{Perceptual similarity (smaller is better) of rendered images with translation and rotational errors in source images}
\label{fig:nol_sensivity}
\end{figure}

\paragraph{Sensitivity to Different Shapes}%
Fig.~\ref{fig:nol_sensivity} provides experimental results that show the sensitivity of NOL with different shapes (cylindrical, box) and pose errors of input images. A synthetic setup is used to ensure precise GT poses and 3D models. For a target pose of an object, we render 5 source images from different viewpoints. 3D models of \textit{cracker\_box} and \textit{mastershef\_can} (in SMOT, we label this object as \textit{maxwell\_can} since it has a different texture) in YCB-V are used to compare the results for different shapes. 50 sets (a set consists of a GT image, a target pose, and 5 source images) are created for each object. We directly measure the image quality using perceptual similarity~~\cite{Zhang_2018_CVPR} between GT and NOL images while increasing the range of pose errors of source images. The results clearly show that the quality of images becomes worse with larger errors. On the other hand, the proposed refinement step successfully reduces the gap between rendered and target images regardless of the shapes of objects.

\section{Conclusion} \label{conclusion}
This paper proposed a novel method that creates training images for pose estimators using a small number of cluttered images. To the best of our knowledge, this is the first attempt to learn multiple objects from a cluttered scene for 6D pose estimation, which minimizes the effort for recognizing a new object. We hope this work highlights the fact that not only estimation methods but also the creation of training images using a few observations are important for real applications. Our code and dataset are publicly available to motivate future research in this direction.

For future work, the method can be extended to optimize 3D models for reducing geometrical errors. This accomplishes the fully self-supervised learning of objects from cluttered scenes in real environments.

\subsubsection*{Acknowledgment}
{The research leading to these results has partially funded by the Austrian Science Fund (FWF) under grant agreement No.~I3969-N30 (InDex) and the Austrian Research Promotion Agency (FFG) under grant agreement No.~879878 (K4R).
}
{
\bibliographystyle{splncs04}
\bibliography{egbib}
}
\newpage
\appendix
\setcounter{figure}{0}
\setcounter{table}{0}    
\renewcommand\thefigure{\thesection.\arabic{figure}}    
\renewcommand\thetable{\thesection.\arabic{table}} 
\section{SMOT Dataset}
\begin{figure}
\centering
\begin{minipage}{.5\textwidth}
\centering
\includegraphics[width=1\linewidth]{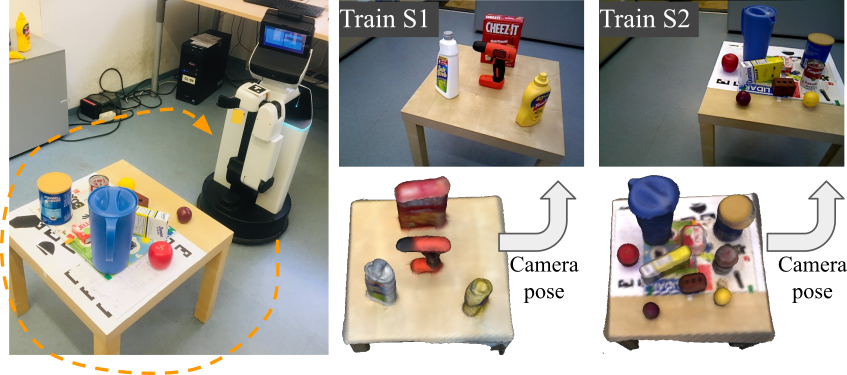}
\caption{Training images of SMOT is collected using a mobile robot driving around the table.}
\label{fig:smot_train}
\end{minipage}%
\begin{minipage}{.5\textwidth}
\centering
\includegraphics[width=0.98\linewidth]{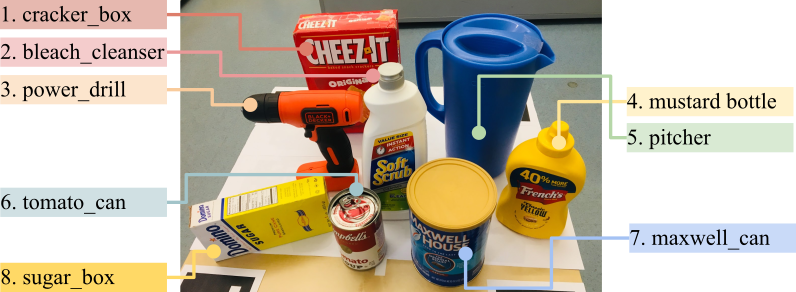}
\caption{Target objects of SMOT.}
\label{fig:smot_objects}
\end{minipage}%
\end{figure}

\begin{figure}
\centering
\includegraphics[width=0.95\linewidth]{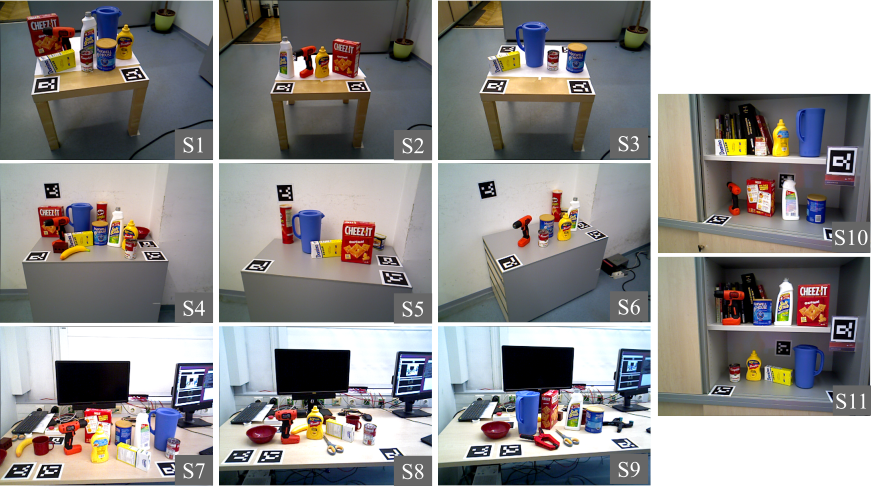}
\caption{Test sequences of SMOT.}
\label{fig:smot_test}
\end{figure}

Fig.~\ref{fig:smot_train} shows 2 training sequences and the reconstruction of each scene. Fig.~\ref{fig:smot_test} lists 11 test scenes of SMOT. All sequences are captured by an Asus Xtion Pro, 640$\times$480 resolution, mounted on the head of a mobile robot. Eight target objects and their names are presented in Fig.~\ref{fig:smot_objects}. Statistics of the dataset are summarized in Table~\ref{table:smot_stat}. 

\setlength{\tabcolsep}{3.0pt}
\begin{table}[hbt]
\begin{center}
\caption{Statistics of SMOT. Training images have a limited elevation range in comparison to the range of test images.}
\label{table:smot_stat}
\begin{tabular}{r|| c| c |c |c | c| c|c| c}

\hline
Object &\begin{tabular}{@{}c@{}}cracker\\box\end{tabular} &  bleach &  driller &  mustard &  pitcher & \begin{tabular}{@{}c@{}}tomato\\can\end{tabular} & \begin{tabular}{@{}c@{}}maxwell\\can\end{tabular}& \begin{tabular}{@{}c@{}}sugar\\box\end{tabular} \\ 
\hline
No. Test images & 2155 & 2171 & 2118 &  2118 &  2090 & 2053 & 2106 & 2037  \\ 
\hline

Train-Azimuth & \multicolumn{4}{c|}{(-180$^\circ$, 180$^\circ$)} &\multicolumn{4}{c}{(-180$^\circ$, 180$^\circ$)}   \\
Train-Elevation& \multicolumn{4}{c|}{(38.3$^\circ$, 40.0$^\circ$)} &  \multicolumn{4}{c}{(38.9$^\circ$, 40.8$^\circ$)}  \\ 
\hline
\hline
Test-Azimuth&  \multicolumn{8}{c}{(-180$^\circ$, 180$^\circ$)}  \\ 
Test-Elevation&  \multicolumn{8}{c}{(8$^\circ$, 42$^\circ$)}  \\ 
\hline
\end{tabular}
\end{center}
\end{table}

\setcounter{figure}{0}
\setcounter{table}{0}  
\section{Implementation Details}

\subsection{Examples of training batches}
\begin{figure}[hbt]
\centering
\includegraphics[width=1.0\linewidth]{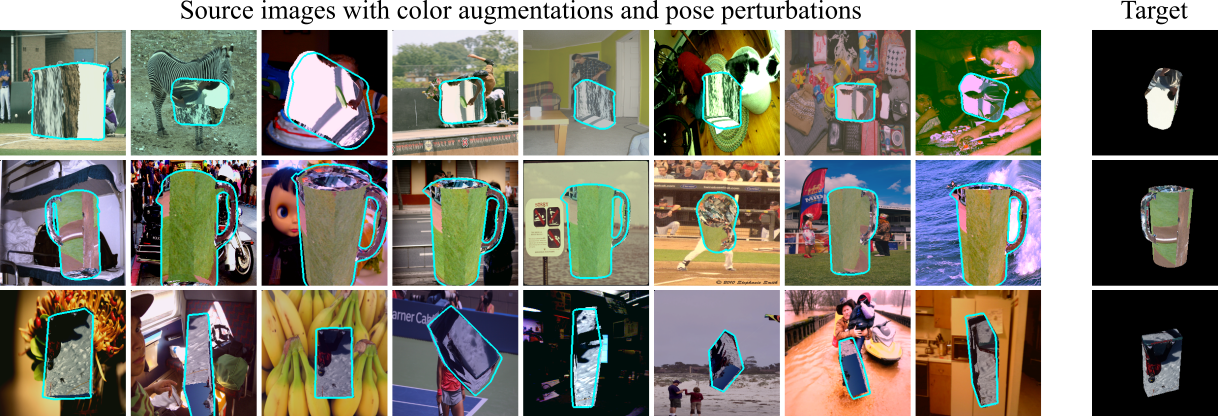}
\caption{Examples of training batches.}
\label{fig:batch}
\end{figure}

Examples of source images and target images are depicted in Fig.~\ref{fig:batch}. Color augmentations are applied to source images and pose perturbations are applied to pose annotations of source images with the parameters in Table~\ref{table:aug_params_}. No augmentation is applied to target images and target poses. 

\setlength{\tabcolsep}{2.5pt}
\begin{table}[hb]

\caption{Parameters of color augmentations and pose perturbations}
\begin{center}
\begin{tabular}{ c| c| c| c|c ||c|c}
\hline
   \multicolumn{5}{c||}{Color augmentation} & \multicolumn{2}{c}{Pose augmentation} \\

  \hline
  Color add & \begin{tabular}{@{}c@{}}Contrast\\norm\end{tabular} & Multiply & \begin{tabular}{@{}c@{}}Gaussian\\blur\end{tabular} & \begin{tabular}{@{}c@{}}Addictive\\noise\end{tabular} &
  \begin{tabular}{@{}c@{}}$\Delta$Trans\\(m)\end{tabular}
   & \begin{tabular}{@{}c@{}}$\Delta$Rot\\(rad)\end{tabular} \\
 
 \hline
 $\mathcal{U}$(-15, 15) & $\mathcal{U}$(0.8, 1.3) & $\mathcal{U}$(0.8, 1.2) & $\mathcal{U}$(0.0, 0.5)& $\mathcal{N}$(0, 10) & $\mathcal{U}$($\pm$ 0.01) & $\mathcal{U}$($\pm$ 0.05)\\
  \hline
\end{tabular}
\end{center}
\label{table:aug_params_}
\end{table}

\subsection{Training}
The Adam optimizer with a learning rate of 0.001 is used and the learning rate is divided by a factor of 10 after 20 epochs during the training of 35 epochs. Both training and evaluation are performed using an Nvidia GTX 1080 with 8Gb memory and i7-6700 CPU. Due to the limitation of our GPU memory, weights are updated after every 10 iterations using average gradient values of the last 10 iterations, which is equivalent to 10 mini batches per iteration. Table~\ref{table:aug_params_} reports ranges of color augmentations and pose perturbations used for training.

\subsection{Network architectures}
\begin{figure}[hbt]
\centering
\includegraphics[width=0.9\linewidth]{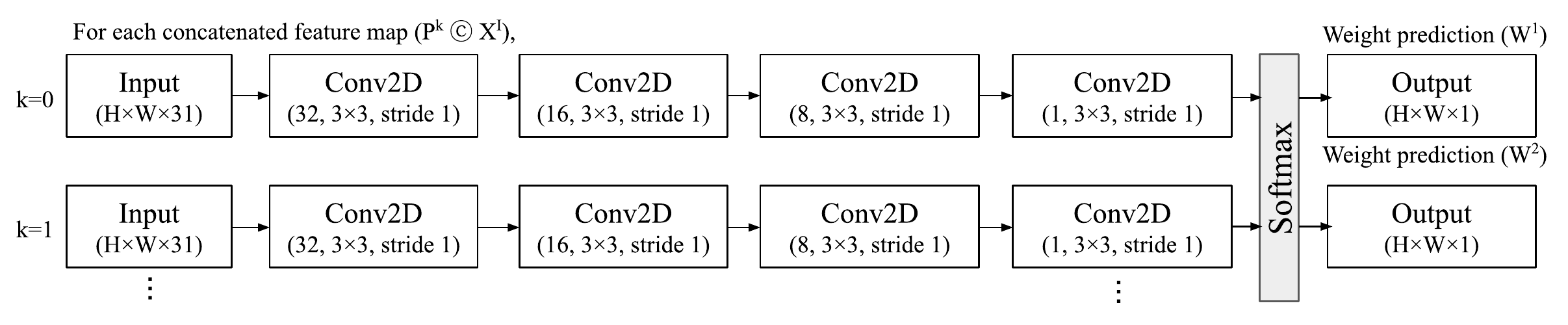}
\caption{The architecture of the weight prediction block.}
\label{fig:weight_pred_block}
\end{figure}

\begin{figure}[hbt]
\centering
\includegraphics[width=0.9\linewidth]{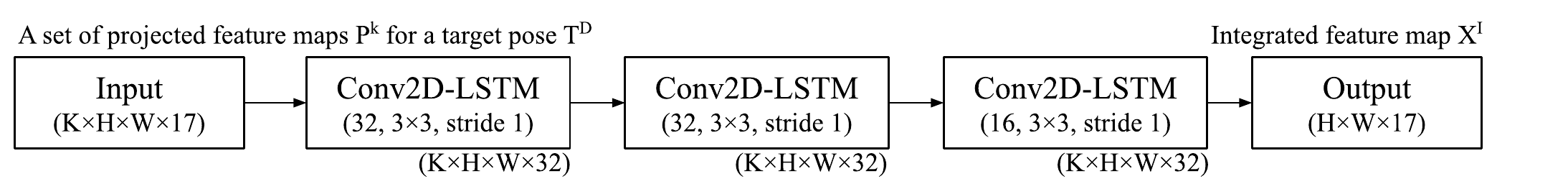}
\caption{The architecture of the LSTM block that integrates projected feature maps.}
\label{fig:lstm_block}
\end{figure}

Fig.~\ref{fig:weight_pred_block} and Fig.~\ref{fig:lstm_block} show architectures of each module in the NOL pipeline.

\setcounter{figure}{0}
\setcounter{table}{0}  
\section{Details of The SMOT Evaluation}%
\begin{figure}[hbt]
\centering
\includegraphics[width=0.9\linewidth]{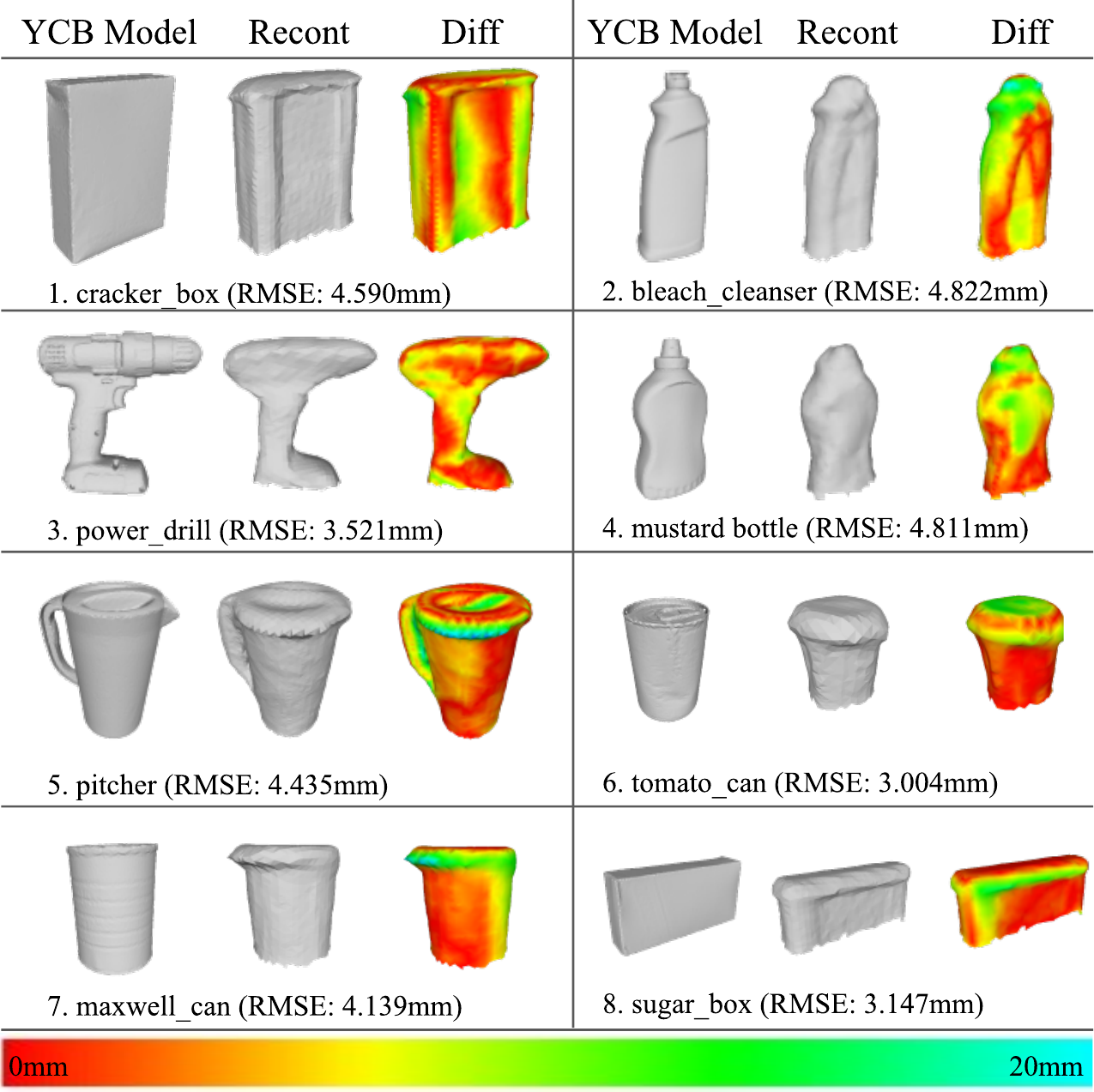}
\caption{Visualization of geometrical errors measured with the Hausdorff distance.}
\label{fig:geo_errors}
\end{figure}

Table~\ref{table:smot_obj} reports object-wise results of the SMOT evaluation. Geometrical errors of reconstructed models used in the SMOT evaluation are visualized in Fig.~\ref{fig:geo_errors}.

\setlength{\tabcolsep}{4pt}
\begin{table}[hbt]
\begin{center}
\caption{Object-wise results of the SMOT evaluation. The ADD metric is used except for symmetric objects, marked with (*), that are evaluated with the ADI metric.}

\label{table:smot_obj}
\begin{tabular}{r|| c c c |c || c  c c |c}
 Type  &  \multicolumn{4}{c||}{RGB} & \multicolumn{4}{c}{RGB-D (ICP3D)} \\
\hline
 3D Model  &  \multicolumn{3}{c|}{Precise} & \multicolumn{1}{c||}{Recont} &  \multicolumn{3}{c|}{Precise} & \multicolumn{1}{c}{Recont}  \\
\hline
Train source  & Real & G2Ltex & \textbf{NOL} & \textbf{NOL} & Real & G2Ltex & \textbf{NOL} & \textbf{NOL}\\
\hline
cracker\_box  & 30.8 & 24.8 &49.5 & 45.4 & 85.2 &92.5 &96.3 &88.6 \\
bleach\_cleanser  & 19.1 & 27.5 &32.7 & 12.8 & 94.0 &89.9 &93.6 &64.9 \\
driller  & 23.8 & 2.3 &19.8 & 26.0 & 87.8 &53.9 &96.4 &91.2 \\
mustard  & 2.0 & 33.2 &25.9 & 19.0 & 88.3 &73.8 &89.7 &82.0 \\
pitcher*  & 25.9 & 21.7 &30.9 & 34.8 & 93.3&92.9 &88.7 &96.1 \\
tomato\_can*  & 36.7 & 17.5 &41.3 & 11.1 & 86.9 &79.0 &84.9 &71.7 \\
maxwell\_can*  & 37.6 & 40.9 &54.7 & 18.3 & 95.3 &94.2 &93.3 &84.6 \\
sugar\_box   & 23.7 & 37.9&29.0 & 12.3 & 60.8 &79.9 &76.9 &55.2 \\

\hline
Average & 25.0 &25.7 &35.5 & 22.5 & 86.5 &82.0 & 90.0 &79.3 \\
\end{tabular}
\end{center}
\end{table}

\end{document}